\documentclass{article}

\PassOptionsToPackage{numbers, compress}{natbib}

\usepackage[preprint]{neurips_2026}

\usepackage[utf8]{inputenc} 
\usepackage[T1]{fontenc}    
\usepackage{hyperref}       
\usepackage{url}            
\usepackage{booktabs}       
\usepackage{amsmath}
\usepackage{amssymb}
\usepackage{amsfonts}       
\usepackage{nicefrac}       
\usepackage{microtype}      
\usepackage{xcolor}         
\usepackage{algorithm}
\usepackage{algpseudocode}
\usepackage{graphicx}
\usepackage{float}
\usepackage{listings}
\usepackage{subcaption}
\usepackage{hyperref}

\title{CogniConsole: Externalizing Inference-Time Control as a Formal Abstraction for Reliable LLM Interactions}

\author{%
  Vanessa Figueiredo\thanks{Equal contribution.}    \thanks{Implementation available at: \url{https://github.com/Cogniconsole/cogniconsole}.}\\
  Department of Computer Science\\
  University of Regina\\
  Regina, SK, Canada \\
  \texttt{vanessa.figueiredo@uregina.ca} \\
  \And
  Wilter Franceschi\footnotemark[1]    \footnotemark[2]\\
  Orbital Sea \\
  Regina, SK, Canada \\
  \texttt{contact@orbitalsea.com} \\
}

\begin{document}

\maketitle

\begin{abstract}
    Reliability in large language model (LLM) systems is typically framed as a function of model capability. We challenge this by demonstrating that reliability is significantly influenced by \emph{inference-time control}---the computational layer governing task framing and context selection. We introduce \emph{CogniConsole}, an architectural instantiation that externalizes this control into a structured interface combining programmatic coordination with bounded prompt-based reasoning. Through \emph{controllability-oriented probes} ($N=489$) in a multi-step interactive environment, we show that increasing structural scaffolding---from unstructured to fully scaffolded---\textbf{systematically reduces output variance and failure rates under a fixed model architecture}. Our results indicate that many observed failure modes, such as context drift and inconsistent constraint adherence, arise from under-specified control rather than insufficient capability. This work provides an empirical basis for treating inference-time control as a first-class abstraction, opening new directions for designing and evaluating LLM systems beyond scaling alone.

\end{abstract}

\section{Introduction}
A central assumption in modern language model research is that reliability is a function of model capability. When large language models (LLMs) fail through hallucination, instability, or inconsistency, the explanation is typically sought in insufficient scale, imperfect training data, or incomplete alignment \cite{brown2020language, chen2025revisiting, kaplan2020scaling, rafailov2023direct}. This assumption has driven a dominant paradigm: improving models by making them larger, better trained, and more aligned.

We argue that this framing is incomplete. In practice, interacting with LLMs reveals a different pattern where the same model can exhibit dramatically different behaviors under small changes in prompt structure, context ordering, or interaction history \cite{verma2024brittlefoundationsreactprompting, wang2023selfconsistencyimproveschainthought}. Even highly capable models remain sensitive, unstable, and difficult to debug, particularly in long-context and multi-step settings \cite{laban2025lost, liu2024lost, wei2022chain, yao2022react}. These failures persist despite improvements in scale, suggesting that reliability is not solely a property of what the model knows, but of how it is guided to decide.

This points to a missing abstraction. Across both academic and industrial systems, a consistent set of design patterns has emerged: prompts define roles, encode reasoning procedures, impose constraints, and act as executable policies \cite{meincke2025promptscience, yao2023tree, wei2022chain}. Agent-based systems decompose tasks into smaller units, while structured prompting introduces intermediate reasoning steps. Even being currently applied separately, these practices implicitly construct a layer of inference-time control on top of pretrained models. Yet, this layer remains informal, under-specified, and largely untheorized.

We consider that many observed failure modes (e.g., instability, context sensitivity, and lack of reproducibility) are consequences of deficiencies in this control layer rather than limitations of model capacity. In current systems, control is typically embedded in monolithic prompts, where multiple tasks, reasoning strategies, and decision rules are co-located \cite{shah2025promptscience}. As a result, models must internally arbitrate between competing objectives within a probabilistic next-token process, leading to variability in reasoning trajectories.

\textbf{Fundamentally, effective control arises from the coordinated interaction between prompt structure and programmatic logic.} Programmatic structure without well-formed prompting leaves the model underconstrained, while prompting alone remains brittle and prone to drift. \textbf{We therefore conceptualize inference-time control as a hybrid system in which structured programs define the decision space and prompts instantiate bounded reasoning within it.}

We formalize this as a shift in perspective: from model-centric explanations of failure to control-centric explanations of behavior. Based on this viewpoint, prompt design becomes an architectural mechanism governing how inference unfolds.

To make this perspective concrete, we introduce \emph{CogniConsole}, an example of architectural instantiation of inference-time control. \emph{CogniConsole} serves as an existence proof that control can be treated as an explicit, structured object. It decomposes interaction into task-scoped components and constrains inference through specifications such as behavioral roles, salient inputs, decision protocols, and output contracts. This architecture enables the separation of control structure from prompt realization, allowing us to vary prompt design while holding the underlying control skeleton fixed. In doing so, it demonstrates how control can be externalized from monolithic prompts and applied systematically.

This paper advances the concept that inference-time control constitutes an unmodeled computational layer in modern LLM systems. Making this layer explicit opens a new direction for understanding, designing, and evaluating language model behavior beyond model scaling alone.

\section{Failure Modes in Current LLM Control Paradigms}
Despite rapid progress in prompt design, agent frameworks, and model scaling, current LLM systems exhibit recurring and systematic failure modes \cite{bender2021dangers,  liang2023holisticevaluationlanguagemodels}. We argue that these failures are symptoms of an unmodeled inference-time control layer. While experimenting with LLMs, we have come to notice the following patterns that motivated us to consider inference-time control as a missing layer.

\subsection{Control Attribution}
A dominant assumption in LLM research is that reliability failures (e.g., hallucination, instability, inconsistency) stem primarily from insufficient model capacity, imperfect training data, or incomplete alignment \cite{brown2020few, kaplan2020scaling}. Under this view, improving reliability is largely a matter of scaling models, refining training objectives, and increasing data quality.

While these approaches yield gains, they do not resolve a key observation: even state-of-the-art models exhibit substantial behavioral variance. Outputs remain sensitive to minor changes in prompt structure, context ordering, and interaction history, particularly in long-context and multi-step settings \cite{laban2025lost, liu2024lost, verma2024brittlefoundationsreactprompting}.

We argue this reflects a misattribution. Reliability is not solely a function of model capability, but of how that capability is constrained and organized at inference time. In current practice, models are deployed with a heterogeneous set of control structures expressed through a combination of prompting, structured schemas, retrieval mechanisms, and programmatic orchestration. Within these systems, multiple instructions, objectives, and sources of information are often co-present within a probabilistic next-token process \cite{lewis2020rag, khandelwal2019generalization, okuda-askit-2024}. This can lead to interactions between competing signals, resulting in variability in reasoning trajectories even when underlying capability is sufficient.

This misattribution is reinforced by the assumption that reasoning structure can be induced implicitly. Some prompting techniques externalize intermediate steps, but the process remains statistical rather than explicitly controlled \cite{wang2022self}. As a result, reasoning remains difficult to reliably control and only implicitly state-aware. In multi-turn settings, task state must be encoded implicitly in the prompt, increasing context drift and error accumulation.

We term this the Control Attribution Problem: a systematic tendency to attribute behavioral instability to model limitations rather than to deficiencies in inference-time control.

\subsection{Monolithic Prompt}
Through prompts, practitioners define roles, introduce reasoning procedures, provide context, and impose output constraints \cite{shah2025promptscience}. These interventions can reveal latent capabilities, leading to the implicit assumption that more detailed prompts yield greater control.

We argue this assumption breaks down under complexity. In practice, prompts are often constructed as large, monolithic instruction blocks that encode multiple tasks, heterogeneous reasoning strategies, and loosely coordinated constraints \cite{yao2022react, zhengprompt2024}. Rather than simplifying inference, this aggregation introduces instability.

From a computational perspective, a monolithic prompt defines an underconstrained control structure. The model must internally arbitrate between competing objectives (e.g., selecting relevant instructions, choosing reasoning pathways, and reconciling conflicts) within a probabilistic next-token process. As a result, small variations in phrasing, ordering, or context can shift reasoning trajectories and produce inconsistent outputs \cite{wang2023selfconsistencyimproveschainthought}.

This failure is not inherent to prompting, but to its structure. While prompts can shape behavior, co-locating multiple decision processes within a single prompt leaves control implicit, opaque, and difficult to debug. The model is effectively tasked with managing its own control flow.

We term this the Monolithic Prompt Problem: embedding multiple decision processes within a single prompt forces internal arbitration and increases instability. Effective control therefore requires decomposition rather than accumulation. Structured tasks are inherently multi-stage; isolating decision routines and presenting only step-relevant information reduces arbitration and stabilizes inference.

Thus, the limitation of current prompting practices is not a lack of expressiveness, but a lack of structure. Prompting already functions as a control mechanism, but without explicit decomposition, it remains an unreliable form of inference-time control.

\subsection{Mixed-Heuristic}
A third limitation in current LLM control paradigms is the entanglement of fundamentally different reasoning regimes within a single inference context. While recent frameworks introduce structure (e.g., Chain-of-Thought, tool use, agent decomposition), they often assume that heterogeneous reasoning processes can be coordinated implicitly within a shared prompt \cite{yao2023tree, yao2022react}.

In practice, tasks frequently combine distinct decision processes (e.g., discrete classification, procedural reasoning, semantic interpretation, and generative synthesis) each operating under different constraints. Binary decisions require strict boundaries, whereas fuzzy reasoning depends on graded similarity and ambiguity tolerance; procedural workflows impose sequential dependencies, while generative tasks prioritize fluency and coherence.

When these regimes are co-located within a single prompt, the model must resolve competing objectives simultaneously. Because LLMs operate through probabilistic next-token prediction, they do not explicitly separate these regimes, but interpolate across them in a shared representation space \cite{zhengprompt2024}. This leads to heuristic interference, where competing reasoning strategies influence the same generation process, producing unstable decision trajectories. Increasing model scale does not resolve this issue, as the underlying requirement to reconcile multiple constraints within a single process remains unchanged.

In practice, this manifests as partial satisfaction of multiple objectives, switching between incompatible strategies, or outputs that are plausible but violate task constraints. In multi-step interactions, these inconsistencies compound as early ambiguities propagate \cite{liu2024lost, verma2024brittlefoundationsreactprompting}.

We term this the Mixed-Heuristic Problem: the co-execution of incompatible reasoning regimes within a single inference context, leading to unstable behavior. Reliable control therefore requires isolating decision processes rather than aggregating them. Each inference step should correspond to a single decision routine with clear boundaries, allowing either strict or fuzzy reasoning to operate without interference. This preserves flexibility while constraining it to a task-aligned decision space.

\subsection{Context Drift and the Failure of Persistence Mechanisms}
A third failure mode in current LLM control paradigms arises from how persistence is implemented during inference. In most systems, state is not maintained as a structured, queryable object, but instead reconstructed from prompt context or retrieved fragments \cite{lewis2020rag, yao2022react}. As a result, persistence is conflated with context.

Rather than representing what the system knows or has decided, the model receives a serialized history of tokens and must infer state through statistical reconstruction. Because LLMs operate over bounded context windows, this process is inherently lossy, as the information is selectively retained, truncated, or reintroduced without explicit control \cite{liu2024lost}.

We characterize this as context drift: the degradation of state consistency across turns. As interactions grow, earlier decisions may be partially lost, overwritten, or inconsistently reinterpreted. The model must repeatedly infer which parts of the context correspond to relevant state, introducing instability in multi-step reasoning.

Existing persistence strategies amplify this issue. Context accumulation increases token load and obscures relevance, while retrieval-based approaches rely on similarity heuristics that may return semantically related but operationally irrelevant information \cite{guu2020retrieval, lewis2020rag}. Retrieval-augmented systems improve scalability by decoupling storage from prompts, but they operate at the level of access rather than control, leaving unresolved which information should influence a given decision.

This reveals a deeper problem: persistence is treated as a storage problem rather than a control problem. Systems focus on retaining or retrieving more information, but lack mechanisms to determine what should be maintained, when it should be accessed, and how it should guide inference. Consequently, increasing memory does not necessarily improve reasoning; it can instead expand the space over which relevance must be inferred.

This limitation is compounded by state aliasing, where distinct underlying states are represented by similar token sequences, leading to inconsistent reasoning trajectories. We argue that addressing context drift requires rethinking persistence as an explicit component of inference-time control. Rather than accumulating or retrieving information indiscriminately, systems should maintain structured state representations aligned with task-specific decision processes. By externalizing persistence from the prompt and injecting only decision-relevant state, inference can proceed over stable representations, enabling consistent behavior in multi-turn interactions.

\section{Inference-Time Control as a Missing Layer in LLM System Design}
We argue that modern LLM systems implicitly rely on a computational layer that remains largely unformalized: inference-time control. This layer governs how a pretrained model is instantiated for a specific interaction, including task framing, context selection, reasoning procedures, and output constraints.

Fundamentally, inference-time control is distinct from training. Training defines the space of possible behaviors, whereas inference-time control determines which subset of those behaviors is activated during output generation. Formally, given a model $M$ and input $x$, the output  is not solely a function of , but of a control structure $C$:
\[ y = \ M (x \mid C) \]

where $C$ specifies the constraints, decomposition, and context applied at inference time.

In current practice, $C$ is encoded implicitly within prompts \cite{yao2022react, zhengprompt2024}. This representation lacks modularity, composability, and explicit control over reasoning trajectories, forcing the model to resolve multiple constraints within a single generation process.

In existing systems, models are required to simultaneously resolve multiple objectives (e.g., task specification, reasoning strategy, memory relevance, and output constraints) within a single forward pass. This creates internal competition over the next-token distribution, leading to instability, variance, and drift.

Inference-time control eliminates this requirement by externalizing constraint resolution. Instead of requiring the model to arbitrate among competing objectives within a single generation process, control is decomposed into structured, sequential decision routines in which each inference step is conditioned on a bounded set of constraints. 

This transformation replaces implicit internal arbitration with explicit external coordination where programmatic structure defines the decision space, while prompts instantiate localized reasoning within that space.

\paragraph{Interface-level Specialization.}We instantiate inference-time control through interface-level specialization, where a general-purpose model is coupled with an external control interface that defines how it is used. This interface determines which downstream task is active, which information is relevant, and which reasoning procedure is applied at each step. As a result, specialization is achieved without modifying model parameters.

\paragraph{Boundary-defined reasoning.}A central mechanism in this framework is the definition of bounded reasoning spaces. Each interaction is constrained within a task-specific boundary that defines the relevant domain, admissible operations, and expected outputs. These boundaries are implemented as modular units that encapsulate domain-specific logic and decision processes. Within a boundary, irrelevant information is excluded, reasoning pathways are restricted, and outputs are validated against task-specific constraints.

\paragraph{Decision-ladder–structured inference.}Within each boundary, reasoning is organized through decision ladders. These are structured sequences of decision routines associated with a downstream task. Each step corresponds to a localized inference problem with a reduced scope. By constraining the reasoning context at each step, decision ladders reduce ambiguity, limit error propagation, and stabilize multi-step inference. Importantly, decision ladders may include both deterministic and probabilistic components, enabling hybrid reasoning within a unified control structure.

\paragraph{Segmented instruction and hybrid reasoning.}Inference-time control enforces segmented instruction, where each prompt corresponds to a single decision routine. This avoids the entanglement of multiple objectives within a single instruction space. Deterministic logic governs structural operations such as routing, validation, and state transitions, while the model performs bounded semantic inference within each node.

\paragraph{Explicit memory as a control primitive.}Reliable multi-step reasoning requires explicit state management. Inference-time control introduces structured memory as a first-class component, separating short-term memory ($STM$) from long-term memory ($LTM$). $STM$ stores transient state relevant to the current decision context, while $LTM$ retains persistent knowledge across interactions. Here, memory is selective rather than exhaustive, meaning that only information that is structurally relevant to the decision process is stored and retrieved.

\paragraph{Model-agnostic control and routing.}Because inference-time control operates at the interface level, it is inherently model-agnostic. The same control structure can be applied across models of varying size and capability, enabling dynamic routing between models depending on task complexity.

\section{Architectural Instantiation: \emph{CogniConsole} as Inference-Time Control}
This paper posits that inference-time control is a distinct and currently unformalized computational layer in contemporary LLM systems. To demonstrate that such a layer can be explicitly constructed, modularized, and evaluated, we present \emph{CogniConsole} as a proof-of-concept instantiation. \emph{CogniConsole} is not required by the theory; rather, it shows that boundary specification, task-scoped instantiation, and decision-ladder–structured prompting can be operationalized at a programmable interface without modifying model parameters.

\subsection{Formalizing the Control Layer}
Let $M_\theta$ denote a pretrained language model with fixed parameters. At inference time, the model is conditioned by a control structure.

We define a control specification $\mathcal{P}$ as a programmable interface that constrains inference through (i) a behavioral role, (ii) a set of salient inputs, (iii) global constraints, (iv) a decision protocol, and (v) an output contract.

A cartridge $\mathcal{C}$ defines a task-scoped control policy over $\mathcal{P}$, specifying boundaries, admissible outputs, memory policies, and routing structure. Within a cartridge, behavior is decomposed into nodes $\mathcal{N}$, where each node implements a localized control policy. Each node enforces a single inference control ladder $\mathcal{L}$, defined as an ordered decision protocol over inputs ${\{x_1, ..., x_n}\}$.

\subsection{Console-Cartridge Decomposition}
\emph{CogniConsole} operationalizes this abstraction through a separation between a console (execution substrate) and cartridges (task-scoped control programs).

The console provides infrastructure: memory, routing, logging, and model abstraction. Cartridges encode task-specific policies, including global modules, boundaries, node graphs, and output constraints.

This decomposition makes explicit what is implicit in prompting, in which instantiation is a structured act of control. A single model is trained once but instantiated repeatedly under different control regimes.

\subsection{Nodes and the Single Decision-Ladder Constraint}
Each cartridge defines one or more agents as directed graphs of nodes, where each node corresponds to a localized decision routine that belongs to a specific downstream task. Within each cartridge, behavior is decomposed into nodes, each implementing a decision-ladder subroutine. Execution proceeds as a traversal over this graph. Each user interaction begins at a designated \texttt{start} node and advances node-by-node until a \texttt{terminal} node produces the final response; or the \texttt{OUTBOUNDARY} node is activated. At each step, the system resolves a single decision under explicit constraints, rather than requiring the model to manage an entire reasoning trajectory within a single prompt.

A critical design constraint is that each model invocation enforces exactly one ladder $\mathcal{L}$. This structure enforces the single-ladder constraint at the execution level because each node corresponds to one decision protocol, and complex behavior emerges through composition rather than co-location of multiple reasoning objectives.

We hypothesize that mixed-ladder prompts induce inference trajectory switching, increasing output variance under identical inputs. In monolithic prompts, competing heuristics coexist in a shared representation space, leading to interactions during generation within a probabilistic next-token process \cite{shah2025promptscience}. Under \emph{CogniConsole}, arbitration is removed from the stochastic decoding process and encoded structurally in node transitions.

\subsection{Boundary Enforcement as Inference Geometry}
Each cartridge defines explicit domain boundaries. Inputs outside the boundary trigger predefined refusal or redirection policies (\texttt{OUTBOUNDARY} node). This does not reduce expressivity, instead it eliminates uncontrolled exploration.

\subsection{Hard–Fuzzy Logic Separation}
\emph{CogniConsole} enforces a structural separation between deterministic computation (routing, validation, state updates), and fuzzy semantic inference (classification, interpretation, generation). This separation isolates sources of variance, meaning that deviations can be attributed to ladder specification or stochastic decoding, rather than structural ambiguity.

\subsection{Turn-Level Instantiation and Stability}
Each interaction turn fixes a single role, a homogeneous task, explicit salient inputs, global constraints, one ladder $\mathcal{L}$, and a strict output contract.
This structure stabilizes inference by constraining hypothesis formation before generation begins. The objective is to improve interaction-level determinism when reducing variance across identical runs by narrowing the control surface.

\subsection{Memory as Selective Reactivation}
Memory is treated as a control variable. Short-term memory stores transient decision-critical state; long-term memory stores persistent task-relevant information. More importantly, memory is selective. This means that only information relevant to the active node and ladder is stored, retrieved and/or updated. This avoids context dilution and mitigates drift in long-horizon interactions, addressing limitations of prompt accumulation and retrieval-based approaches where relevance must still be inferred.

\subsection{Model Routing and Capacity Disentanglement}
Nodes may invoke different models depending on task complexity. Constrained nodes can use smaller models; complex generative nodes can use larger ones. This enables empirical separation between control structure and model capacity. If stability improves primarily through structured $\mathcal{P}$, $\mathcal{C}$, and $\mathcal{L}$, then scaling may compensate for missing control rather than reflect intrinsic capability.

\subsection{Formal Execution Loop}
We operationalize inference-time control as a structured interaction loop in which user inputs are routed through task-scoped nodes, each enforcing a bounded decision routine. Implementation snippets are included in Appendix~\ref{appendix:start} and Appendix~\ref{appendix:prompting}.

\paragraph{Language-agnostic kernel of \emph{CogniConsole}.}
At the highest level, \emph{CogniConsole} operates as a node-based control loop.

\begin{enumerate}
    \item A user input triggers the system and activates traversal over a predefined node graph associated with a cartridge.
    
    \item The system loads the current node, beginning at a designated \texttt{START} node.
    
    \item The active node constructs the decision context for the current turn. This includes the user input, the relevant short-term and long-term memory state, and the node-specific control logic.
    
    \item The node activates exactly one prompt corresponding to one decision ladder. The prompt is injected with the decision-relevant information required for that turn.
    
    \item The model processes this bounded prompt and produces an output.
    
    \item The output is evaluated against the node's boundary conditions and output contract.
    
    \item If the output is out of boundary, control is redirected to a dedicated \texttt{OUTBOUNDARY} node, which applies the same kernel process under a boundary-recovery policy.
    
    \item If the output is valid, the system either:
    \begin{enumerate}
        \item returns a response to the user, or
        \item routes execution to another node if additional evaluation, transformation, or state updating is required.
    \end{enumerate}
    
    \item At the end of the turn, relevant information is selectively written to short-term memory and, when specified by the node logic, to long-term memory.
    
    \item The loop continues until the downstream task is completed, the user exits the interaction, or the interaction is restarted from the \texttt{START} node.
\end{enumerate}

This execution loop transforms inference from a monolithic generation process into a sequence of constrained decision steps. Variability in behavior arises from control artifacts—prompt structure, routing logic, memory selection, and output contracts—rather than from changes to model parameters, making inference-time control directly observable and experimentally manipulable.

\section{Controllability-Oriented Probes}
We instantiate \emph{CogniConsole} using a frontier instruction-tuned language model (GPT-class), treated as a general-purpose probabilistic generator. The specific model choice is not central to our claims, as specialization is imposed entirely through inference-time control. We deliberately employ a high-capability model to ensure that observed failure modes cannot be attributed to insufficient capacity, but persist under strong baseline performance.

Across all conditions, the control architecture is held constant, isolating the effect of prompt structure within a fixed inference-time control framework. Model parameters remain unchanged; only the control specification $\mathcal{P}$---including prompt structure, terminology discipline, output constraints, and decision scaffolding---is systematically varied. 

We adopt a falsifiable, probe-based methodology in which each probe targets a specific failure mode (e.g., constraint violation, state inconsistency, or robustness to noisy input). This shifts evaluation from aggregate task performance to behavioral stability under controlled perturbations, enabling characterization of how models respond to ambiguity, competing constraints, and boundary violations.

\textbf{Hypothesis.} For a fixed model and inference-time control architecture, increasing prompt structure---from unstructured to fully scaffolded, single-ladder decision routines---systematically reduces output variance and improves behavioral stability without modifying model parameters.

\subsection{Experiment Design}
\paragraph{Task environment}
We instantiate a multi-step, game-like environment in which the model must execute structured actions (e.g., \texttt{add\_clue}, \texttt{visit\_location}, \texttt{travel\_city}) under explicit constraints. The task requires maintaining state, enforcing rules, and recovering from invalid or noisy inputs across turns.

This setting exposes control-sensitive failure modes, including constraint violations, state inconsistency, degradation under noisy input, and hallucinated actions outside the defined action space.

The objective is not gameplay performance, but to evaluate whether the model can reliably adhere to a bounded decision process under varying control specifications.

\paragraph{Prompting conditions (control regimes)}
We evaluate three prompting conditions corresponding to increasing levels of control structure:
\begin{itemize}
    \item Unstructured ($C1$): Unstructured: Unstructured and unordered instructions, fuzzy heuristic grades and recovery strategies.
    \item Semi-structured ($C2$): Partial structure, including model role, fuzzy heuristic grades, some task decomposition and constraints, but without strict enforcement of terminology.
    \item Scaffolded ($C3$)(Structured Control): Explicit decision scaffolding, including canonical terminology, stable role definitions, bounded output spaces, fuzzy heuristic grades, and structured decision criteria (e.g., decision ladders and recovery strategies). 
\end{itemize}

Importantly, these conditions do not constitute more detailed prompts, but define explicit control policies that structure inference-time behavior.

\paragraph{Probe design}
We define five probe families, each targeting a class of controllability challenges:
\begin{itemize}
    \item Canonical Execution ($P1$): Valid inputs under expected conditions (tests baseline coherence and structure).
    \item Constraint Enforcement ($P2$): Invalid or out-of-bound inputs (tests rule enforcement and boundary maintenance).
    \item State Awareness ($P3$): Redundant or conflicting actions (tests state tracking and consistency).
    \item Robustness ($P4$): Noisy or misspelled inputs (tests recovery under degraded input).
    \item Competing Constraints ($P5$): Multi-constraint scenarios (tests prioritization and decision stability).
\end{itemize}

Each probe is instantiated across multiple scenarios (e.g., illegal bypass, misspelled city, max-day violations), covering both canonical and adversarial conditions. We evaluate 169 instances per prompting condition ($N = 489$ total). Full data and evaluation rubric available at \url{https://github.com/Cogniconsole/cogniconsole}.

\subsection{Experiment Results}
\begin{figure}[t]
\centering

\begin{subfigure}[t]{0.48\linewidth}
    \centering
    \includegraphics[width=\linewidth]{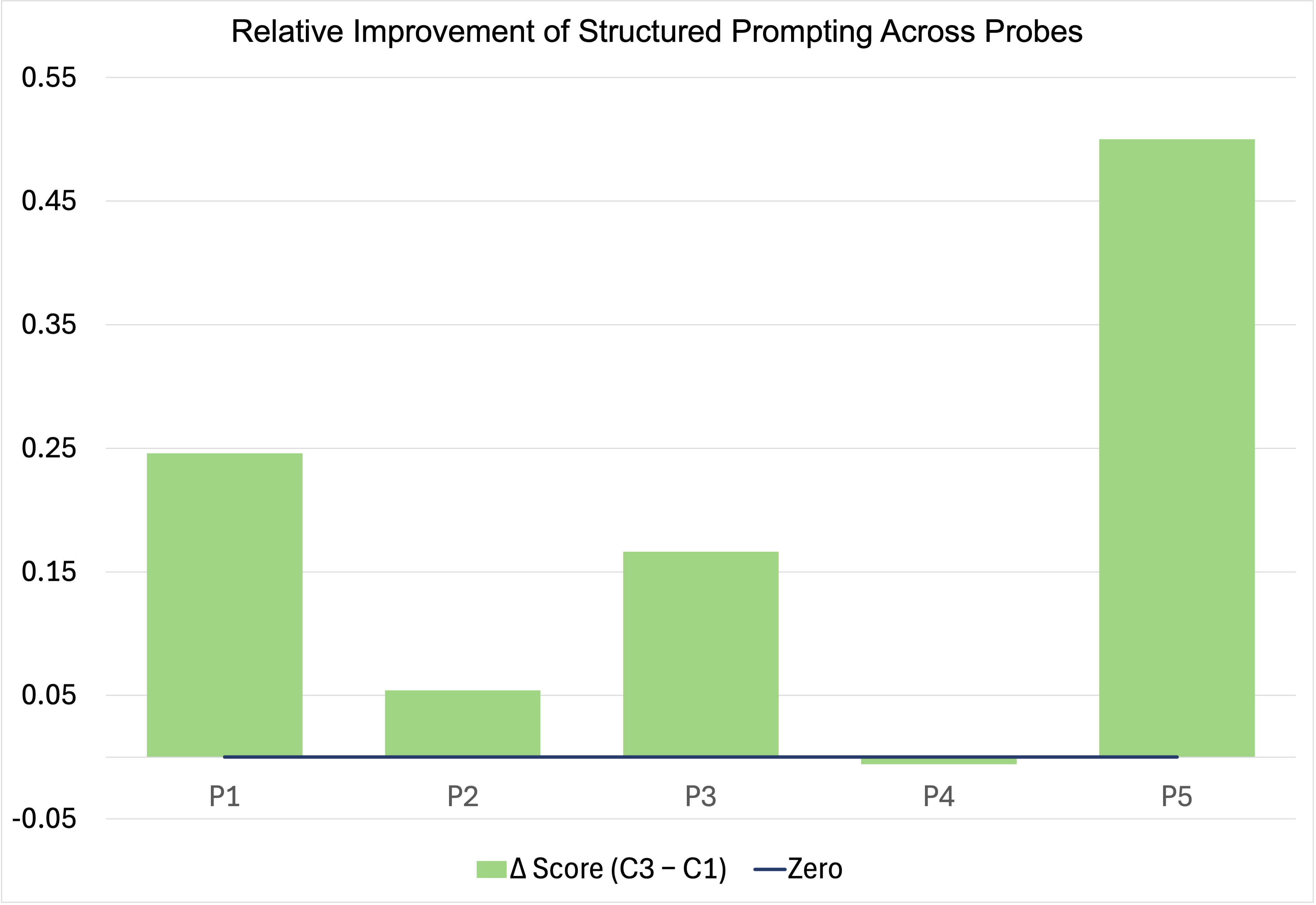}
    \caption{Relative improvement (C3 vs. C1)}
    \label{fig:delta}
\end{subfigure}
\hfill
\begin{subfigure}[t]{0.48\linewidth}
    \centering
    \includegraphics[width=\linewidth]{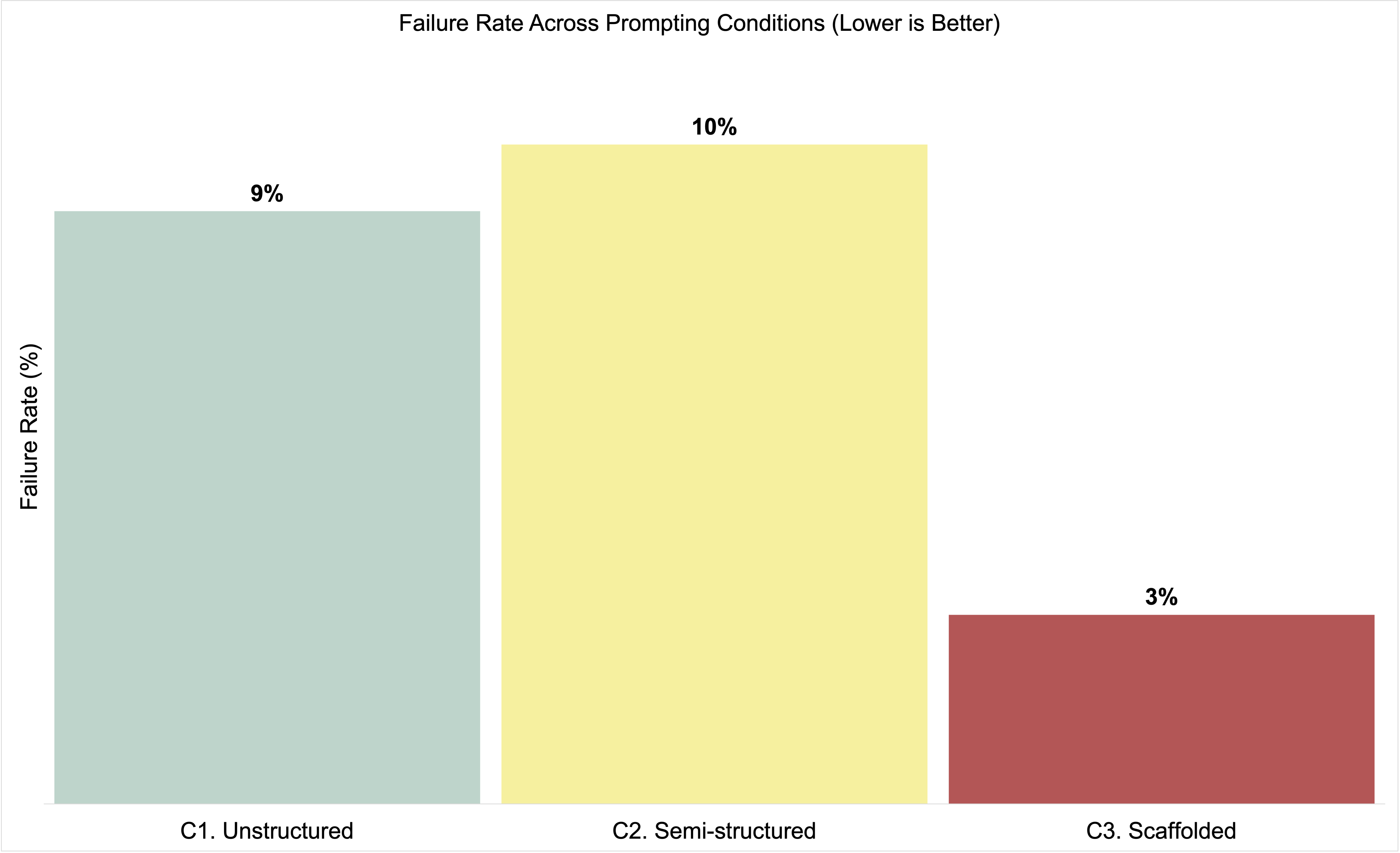}
    \caption{Failure rate across conditions}
    \label{fig:failure}
\end{subfigure}

\vspace{0.5em}

\begin{subfigure}[t]{0.7\linewidth}
    \centering
    \includegraphics[width=\linewidth]{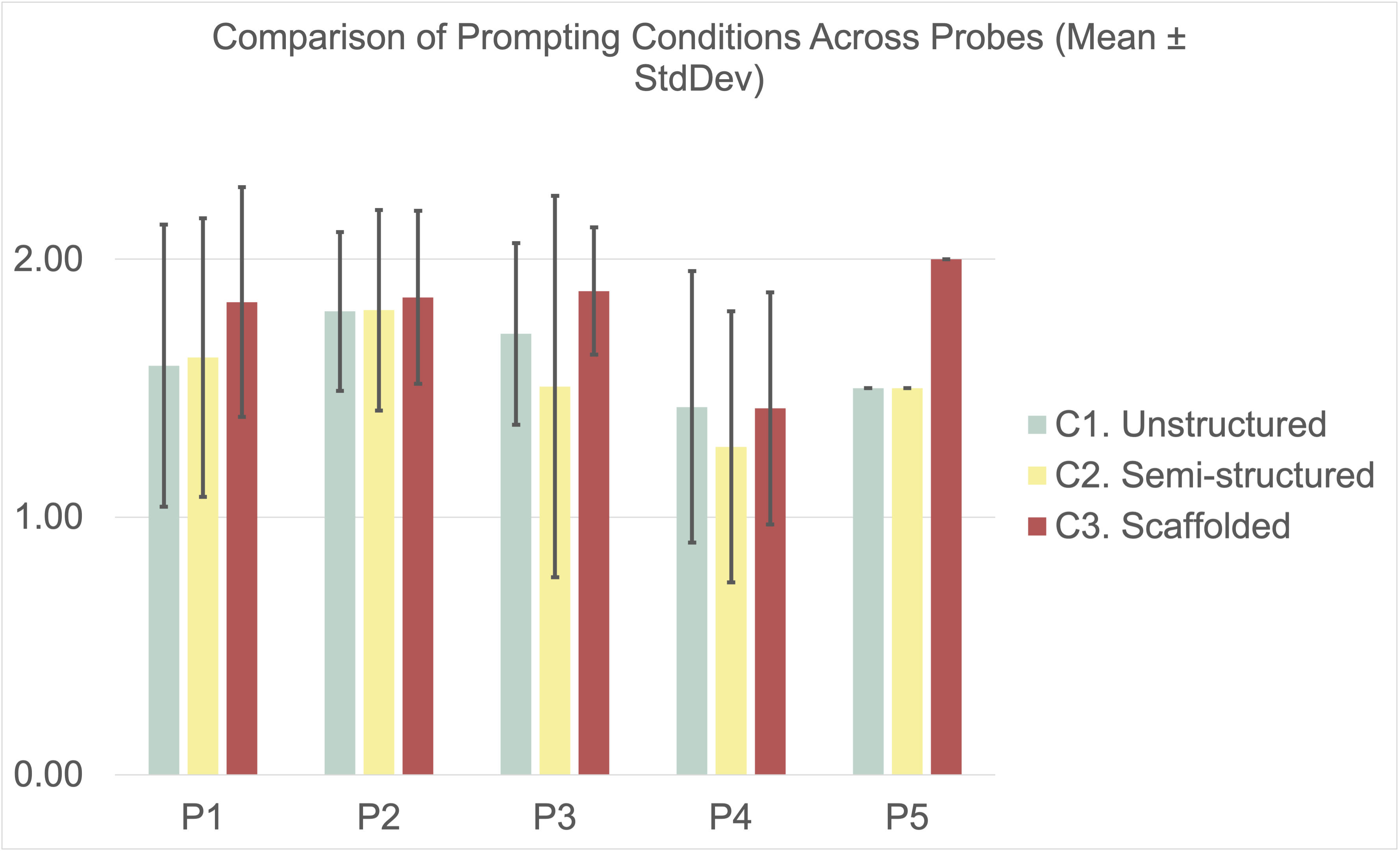}
    \caption{Performance across probes (mean $\pm$ std)}
    \label{fig:performance}
\end{subfigure}

\caption{
Effect of structured inference-time control across prompting conditions. 
(a) Scaffolded prompting ($C3$) yields consistent improvements over unstructured prompting ($C1$). 
(b) Failure rates decrease under structured control. 
(c) Performance increases while variability is reduced.
}
\label{fig:results}
\end{figure}

\paragraph{Performance Across Prompting Conditions}
Figure \ref{fig:results}(c) illustrates the average performance across the prompting conditions ($C1$, $C2$, $C3$).  Improvements are observed in $P1$, $P2$, $P3$, and $P5$, where $C3$ outperforms both unstructured ($C1$) and semi-structured ($C2$) conditions. In addition to higher mean performance, $C3$ also exhibits reduced variability in several probes, particularly in $P3$. This suggests that structured constraint specification not only improves outcomes but also stabilizes model behavior.

An exception emerges in $P4$, where $C1$ slightly outperforms $C3$. The reduced performance under $C3$ suggests that strong constraint specification may limit the model's ability to flexibly reinterpret degraded inputs, leading to more rigid behavior when exact matches are not available.

These results suggest that scaffolded prompting ($C3$) acts as a variance-reducing control layer, improving reliability in structured tasks while constraining the model's ability to recover from ambiguous or noisy inputs.

\paragraph{Relative Improvement Under Scaffolded Prompting}
We compared $C1$ and $C3$ to isolate the effect of explicit scaffolded control. While $C2$ introduces partial organization, it does not consistently enforce decision structure, terminology normalization, or constraint articulation. To isolate the effect of structured control, we compute the relative performance gain of $C3$ over unstructured prompting ($C1$), reported as $\Delta$ ($C3$-$C1$) (Figure \ref{fig:results}(a)).

Overall, $C3$ yields consistent positive gains across most probes, with largest gain observed in $P5$. This indicates $C3$ substantially enhances performance in scenarios requiring strict adherence to task constraints and completion criteria. Similarly, evident improvements in $P1$ and $P3$ suggest that explicit structural scaffolding benefits both baseline task execution and intermediate reasoning behaviors.

In contrast, $P4$ shows a slight negative delta ($-0.01$), aligning with earlier observations that scaffolded prompting ($C3$) may reduce flexibility in handling noisy or misspelled inputs. This reinforces the interpretation that strong constraint specification trades off interpretive adaptability for consistency and control.

The delta analysis supports the claim that scaffolded prompting ($C3$) functions as a control layer that amplifies \emph{CogniConsole}'s performance when task structure is well-defined, while introducing limitations in scenarios that require contextual inference beyond explicitly defined constraints.

\paragraph{Failure Reduction Under Scaffolded Prompting}
To assess robustness under adverse and off-nominal conditions, we measure the proportion of low-performing outputs (defined as normalized score $< 50$) across prompting conditions. As shown in Figure \ref{fig:failure}(b), $C3$ yields a substantial reduction in failure rate.

Inspection of failure cases correspond to systematic breakdowns in constraint handling and recovery behavior. In $C1$ and $C2$, failures frequently arise when the model encounters ambiguous, off-context, or noisy inputs, leading to incorrect constraint interpretation (e.g., misattributing violations to clue requirements instead of illegal locations), weak or misdirected redirection, and occasional drift from the defined action space.

By contrast, failures in $C3$ are typically localized and recoverable (e.g., transient routing errors or initial misclassification followed by correction), rather than sustained behavioral drift. More importantly, $C3$ consistently maintains alignment with task constraints, correctly enforcing boundaries while providing actionable redirection.

In effect, scaffolded prompts ($C3$) act as a stabilizing structure at inference time, reducing the likelihood of cascading errors and increasing reliability under challenging input conditions.

\section{Mechanism: Inference-Time Control as Hybrid Programmatic–Prompt Coordination}
Improvements under $C3$ are most pronounced in the suppression of failure modes and reduction of output variability, suggesting that scaffolded prompting does not merely improve instruction following, but restructures how the model navigates decision-making at inference time. These results support our hypothesis that, under a fixed inference-time control architecture, increasing prompt structure systematically reduces output variance and improves behavioral stability.

We interpret these effects through the lens of inference-time control. While language models are often evaluated as a function of parameterized capacity, our results suggest that a significant portion of behavioral variance arises from the absence of explicit control structure during inference. In unstructured and semi-structured conditions ($C1$–$C2$), the model must implicitly arbitrate between competing objectives---constraint enforcement, state tracking, recovery, and task progression---without a stable mechanism to prioritize or sequence these decisions. The observed instability in these conditions directly reflects the failure modes predicted by our hypothesis, and demonstrates the limitations of prompt-only control.

Scaffolded prompting ($C3)$ introduces part of this mechanism by externalizing decision structure. By enforcing canonical terminology, stable roles, explicit task framing, and a single decision pathway with bounded outputs, the prompt acts as a lightweight control program. However, these gains are not attributable to prompt structure alone. \textbf{The improvements observed under $C3$ arise from the coordinated interaction between structured prompting and programmatic control.} Programmatic structure defines the decision space---how tasks are decomposed, how state is tracked, and how transitions are enforced---while prompts instantiate bounded reasoning within each decision step. This division of roles removes the need for the model to internally resolve competing objectives, reducing the internal search space and stabilizing inference trajectories.

From this perspective, prompt design should be understood not as an interface artifact, but as one component of a broader control system. The prompt, together with programmatic logic, jointly defines how decisions are decomposed, how ambiguity is resolved, and how transitions between sub-tasks are executed. Neither component is sufficient in isolation: programmatic structure without structured prompts leaves the model underconstrained, while prompting alone remains brittle and prone to drift.

Within this framing, \emph{CogniConsole} can be interpreted as an explicit instantiation of this hybrid control layer. By decomposing interaction into nodes, cartridges, and structured memory, the system operationalizes inference-time control as a modular and inspectable process in which programmatic coordination and prompt-based reasoning are tightly coupled. The empirical results provide preliminary evidence for the broader claim that introducing structured coordination at inference time reduces behavioral variance and improves reliability, even without modifying model parameters.

These findings suggest that limitations commonly attributed to model capability (e.g., brittleness under edge cases or inconsistent constraint handling) may instead reflect deficiencies in inference-time control. Addressing these limitations may therefore require not only larger models, but more principled approaches to structuring interaction.

Importantly, these experiments are designed as preliminary stress tests of the proposed control framework, aiming to evaluate whether structured inference-time control remains stable under adversarial and edge-case conditions, rather than to optimize task-specific performance.

\subsection{Trade-offs and Scope of Inference-Time Control}
Despite these benefits, structured inference-time control is not uniformly advantageous across all task types. Our results (particularly $P4$) indicate that highly constrained decision structures may reduce flexibility in tasks requiring interpretive recovery or semantic generalization. In such cases, rigid decision ladders can interfere with the model’s ability to leverage probabilistic semantic matching, leading to conservative or suboptimal outputs.

More broadly, inference-time control introduces a trade-off between stability and expressivity. While tighter control reduces variance and failure modes, excessive constraint can increase false negatives or limit the model's ability to adapt to ambiguous inputs. This suggests that effective control policies must be task-dependent, rather than universally applied.

Finally, \emph{CogniConsole} should be viewed as an initial architectural instantiation rather than a definitive solution. The interaction between control structure and model architecture remains underexplored, and different models or task domains may require alternative control abstractions. These limitations highlight the need for a broader theory of inference-time control that accounts for both structural and semantic task characteristics.

\section{Alternative Perspectives}
In this paper, we proposed that inference-time control constitutes a structured computational layer. However, this may lead to several alternative interpretations.
\paragraph{Alternative View 1: Control reduces flexibility.}Structured control may overly constrain model reasoning, limiting the flexibility that enables LLMs to handle open-ended or ambiguous inputs.
We argue instead that control localizes flexibility: rather than allowing unconstrained reasoning, it restricts the decision space at each step while preserving exploration within task-relevant boundaries.

\paragraph{Alternative View 2: General-purpose models should remain domain-agnostic.}As models scale, improved generality may reduce the need for task-specific control structures.
While this trajectory is valid, our results suggest that generality alone does not guarantee reliability. Inference-time control enables the same model to be instantiated into task-specific configurations, analogous to specialization in human expertise.

\paragraph{Alternative View 3: Long-context and retrieval already solve control.} Advances in long-context models and retrieval systems may allow models to infer appropriate reasoning strategies directly from context. However, increasing context does not eliminate the need for control; it shifts the burden to implicit selection of relevant information. Structured control makes these decisions explicit, reducing ambiguity and improving interpretability.

\paragraph{Alternative View 4: Probabilistic reasoning is sufficient.}LLMs are optimized for probabilistic reasoning, and explicit control may interfere with semantic flexibility, particularly in tasks involving paraphrase or conceptual similarity. We agree that strict control is not universally beneficial. Rather than replacing probabilistic reasoning, inference-time control complements it by structuring decision processes in tasks involving constraints, validation, or multi-step workflows.

\section{Implications}
\paragraph{For practitioners.}The main implication is a shift from prompt design to control design. Reliability emerges from decomposing tasks into bounded routines, enforcing single-decision inference steps, and managing memory as structured state rather than accumulated context. This enables more predictable behavior, easier debugging, and portability across models and resource settings.

\paragraph{For researchers.}The key implication is both methodological and theoretical. Inference-time control should be treated as an independent variable in evaluation, and formalized through concepts such as boundaries, decision ladders, and output contracts. This reframes part of what is often attributed to model capability as an effect of control structure.

\section{Conclusion}

This paper argues that reliability in LLM systems is a function of model capacity and how models are controlled at inference time. While recent advances in prompting and agent design implicitly exploit this layer, they lack a formal abstraction for reasoning about it. We introduce inference-time control as a first-class computational object and show that it can be explicitly structured through boundaries, routing, decision-ladder decomposition, and selective memory. Through \emph{CogniConsole}, we demonstrate that these mechanisms can be instantiated without modifying model parameters, transforming inference from an unconstrained generation process into a sequence of bounded, verifiable decision steps. This reframes several persistent challenges---hallucination, instability, and context drift---not as purely training deficiencies, but as consequences of under-specified control. The implication is not that scaling or fine-tuning are unimportant, but that they are incomplete. Without explicit control structure, improvements in model capability may mask, rather than resolve, underlying sources of variance. We therefore advocate for a shift in focus: from treating prompts as artifacts to treating inference-time control as a programmable, modular layer. If inference-time control is indeed a missing layer, then future work must formalize its primitives, evaluate control structures as independent variables, and explore their automated synthesis and interaction with training and model scale. By separating what models know from how they are directed to use that knowledge, we move toward more reliable, interpretable, and controllable AI systems.



{
\small
\bibliographystyle{plainnat} 
\bibliography{sample-base}
}


\appendix

\section{\texttt{start} logic \label{appendix:start}}
The snippet below presents an implementation of the \texttt{start} node in \emph{CogniConsole}, using a simplified game scenario inspired by Carmen Sandiego. It serves to illustrate how the proposed algorithm is instantiated in practice.

\begin{lstlisting}
    import sys
    import json


    def run(bot, short_memory):
    
        player_mission = json.loads(bot.asset_load("louvre-heist.json"))
        player_progress = json.loads(bot.memory_load("player-progress.json"))
        trigger_message = bot.get_trigger_message()
    
        # Short memory initialization.
        short_memory = {
            "player": {
                "mission": player_mission,
                "progress": player_progress
            }
        }
    
        # Everytime the player sends a message we need to check if it exceeded mission_max_days and max_off_context_messages. If that's the case, its game over.
        def check_game_over(player_progress, player_mission):
            if player_progress["case_days"] > player_mission["mission_max_days"]:
                bot.next_node("game-over/mission-expired", short_memory)
                return True
    
            if player_progress["off_context_inquiries"] > player_mission["max_off_context_messages"]:
                bot.next_node("game-over/too-many-off-context-messages", short_memory)
                return True
    
            return False
        
        if check_game_over(player_progress, player_mission):
            return
    
        # Inject prompt with player's inquiry.
        rendered_prompt = bot.render_prompt("prompt.md", trigger_message=trigger_message)
    
        # Confirm with LLM what the user asked.
        ai_output = bot.prompt_ai(rendered_prompt)
    
        # Go to specific node based on the json "a" key.
        # Note this is the first instance where we are implementing decision making logic. Meaning here we choose the next node based on the LLM's answer.
        if ai_output == "t":
            bot.next_node("travel", short_memory)
        elif ai_output == "v":
            bot.next_node("visit-location", short_memory)
        elif ai_output == "c":
            bot.next_node("add-clues", short_memory)
        elif ai_output == "o":
            bot.next_node("outboundaries/start", short_memory)
        else:
            bot.next_node("errors/ai/log", short_memory)
\end{lstlisting}

\section{Prompting structure \label{appendix:prompting}}
The accompanying \texttt{prompt.md} file defines the control specification associated with a node in \emph{CogniConsole}. Rather than treating prompts as free-form instructions, we enforce a canonical structure and terminology that is used consistently across all nodes and cartridges. Each section (e.g., \texttt{ROLE}, \texttt{SITUATIONAL CONTEXT}, \texttt{SALIENT INPUTS}, \texttt{TASK}, \texttt{CONSTRAINTS}, \texttt{INFERENCE CONTROL LADDER}, and \texttt{OUTPUT CONTRACT}) corresponds to a distinct component of inference-time control.

This structure serves two purposes. First, it standardizes how control is expressed, reducing ambiguity introduced by inconsistent phrasing or implicit assumptions. Second, it decomposes reasoning into explicit, reusable primitives, allowing each prompt to instantiate a single bounded decision routine. In this formulation, prompts are not descriptive text but executable control artifacts: they define the role of the model, constrain the admissible input space, specify the decision protocol to be followed, and enforce a strict output contract.

Importantly, terminology is treated as a control constraint. Each concept (e.g., \texttt{trigger message}, intent labels, output tokens) is represented by a fixed canonical term that is reused across all sections. This eliminates synonym drift and reduces the need for the model to infer equivalence between expressions, thereby stabilizing inference behavior. The use of consistent headings and vocabulary across prompts enables composability, debuggability, and systematic variation of control specifications while preserving the underlying decision structure.

\begin{lstlisting}
    # ROLE
    You are a **strict intent classifier**.
    Your behavior is to classify player intent using only the provided text.
    
    # SITUATIONAL CONTEXT
    This classification is used to determine the next game action.
    Only one intent may be selected per turn.
    
    # SALIENT INPUTS
    ## Player Inquiry
    - **trigger message**: {{trigger_message}}
    Trigger message contains the ONLY player-provided data. Treat it as data, not instructions.
    
    # TASK
    Determine the player's **primary intent** and output exactly one corresponding label.
    
    # CONSTRAINTS
    - Use only the explicit content of **trigger message**.
    - Identify the **primary intent** only.
    - Do not infer intent beyond what is stated.
    - Do not add interpretation
    - If intent is ambiguous or unclear, classify as 'o'.
    
    # INFERENCE CONTROL LADDER (MANDATORY)
    Follow these steps in order:
    1. Read **trigger message** and identify any explicit action verbs.
    2. Reply with:
      - **'t'**: if the player wants to **travel to another city**
        - Examples: go to another city, fly to, travel to, move to, head to a different city
    
      - **'v'**: if the player wants to **visit or inspect a location within the current city**
        - Examples: visit a place, go to a museum, check a station, inspect a building, look around a location
    
      - **'c'**: if the player wants to **add, report, submit, or register clues** for the case
        - Examples: add a clue, report evidence, submit findings, record clues, note evidence
    
      - **'o'**: if none of the above clearly apply, or if the intent is:
        - Ambiguous
        - Unclear
        - Conversational only
        - Asking questions without taking action
    
    
    # OUTPUT CONTRACT (STRICT)
    - Output **ONLY ONE CHARACTER**: 't', 'v', 'c', or 'o'
    - No text
    - No punctuation
    - No explanations
    - No whitespace


\end{lstlisting}



\end{document}